\documentclass[11pt]{article}
\usepackage[utf8]{inputenc}
\usepackage[T1]{fontenc}

\usepackage[margin=1in]{geometry}
\usepackage{graphicx}
\usepackage{enumitem}
\usepackage{newunicodechar}
\newunicodechar{−}{\ensuremath{-}}
\newunicodechar{∈}{\ensuremath{\in}}
\newunicodechar{σ}{\ensuremath{\sigma}}
\newunicodechar{Σ}{\ensuremath{\Sigma}}
\newunicodechar{⊙}{\ensuremath{\odot}}
\newunicodechar{≈}{\ensuremath{\approx}}
\newunicodechar{₀}{\ensuremath{_0}}
\newunicodechar{₃}{\ensuremath{_3}}
\newunicodechar{ₖ}{\ensuremath{_k}}
\newunicodechar{ⁿ}{\ensuremath{^n}}
\newunicodechar{×}{\ensuremath{\times}}
\newunicodechar{≥}{\ensuremath{\geq}}
\newunicodechar{≤}{\ensuremath{\leq}}
\newunicodechar{±}{\ensuremath{\pm}}
\newunicodechar{→}{\ensuremath{\rightarrow}}
\newunicodechar{↑}{\ensuremath{\uparrow}}
\newunicodechar{α}{\ensuremath{\alpha}}
\newunicodechar{β}{\ensuremath{\beta}}
\newunicodechar{γ}{\ensuremath{\gamma}}
\newunicodechar{λ}{\ensuremath{\lambda}}
\newunicodechar{μ}{\ensuremath{\mu}}
\newunicodechar{κ}{\ensuremath{\kappa}}
\newunicodechar{δ}{\ensuremath{\delta}}
\newunicodechar{θ}{\ensuremath{\theta}}
\newunicodechar{∼}{\ensuremath{\sim}}
\newunicodechar{≠}{\ensuremath{\neq}}
\newunicodechar{₁}{\ensuremath{_1}}
\newunicodechar{₂}{\ensuremath{_2}}
\newunicodechar{ₙ}{\ensuremath{_n}}
\newunicodechar{⁻}{\ensuremath{^-}}
\newunicodechar{⁴}{\ensuremath{^4}}
\newunicodechar{ᵀ}{\ensuremath{^{T}}}
\newunicodechar{₄}{\ensuremath{_4}}
\newunicodechar{²}{\ensuremath{^2}}
\newunicodechar{¹}{\ensuremath{^1}}

\usepackage{calc}
\makeatletter
\newcommand{\real}[1]{#1}
\makeatother

\usepackage{amsmath,amssymb}
\usepackage{booktabs}
\usepackage{longtable}
\usepackage{array}
\usepackage{caption}
\usepackage{float}
\usepackage[hidelinks]{hyperref}

\title{\textbf{Dual-Resolution Attention-Gated Deep Learning with Ordinal Regression for Diabetic Retinopathy Grading: A Quantified Assessment of Cross-Domain Generalization}}
\author{Afshan Hashmi\\ Tuwaiq Academy, Tuwaiq Research and Development Centre, Riyadh, Saudi Arabia\\ \texttt{afshanhashmi786@gmail.com} \quad ORCID: 0009-0004-4951-8350}
\date{}

\begin{document}
\maketitle

\begin{abstract}
\textbf{Purpose:} Automated diabetic retinopathy (DR) grading is widely proposed to extend screening capacity, but most reported models are validated only on the dataset they were trained on, leaving their behaviour under real screening variability unmeasured. This study quantifies how far grading performance falls when the imaging domain shifts, and what that means for screening.

\textbf{Methods:} A dual-resolution model pairs two EfficientNet backbones: B0 on Ben Graham-normalised input at 224$\times$224 and B3 on CLAHE-enhanced input at 300$\times$300, fused by a learnable attention gate, with an ordinal decomposition head modelling severity as an ordered scale. Training combined 4,149 images (APTOS 2019, n~=~2,929; a Messidor-2 training portion, n~=~1,220); evaluation used a held-out APTOS split (n~=~733) and a Messidor-2 test set (n~=~524) excluded from training and model selection. Performance was assessed with bootstrap confidence intervals and a multi-seed ablation.

\textbf{Results:} Quadratic weighted kappa was 0.882 (95\% CI 0.853-0.906) on APTOS and 0.679 (95\% CI 0.613-0.735) on Messidor-2 for this run, a significant gap of 0.202; across three seeds the held-out kappa was 0.689 $\pm$ 0.021. Referable-DR sensitivity fell from 0.879 to 0.620, below the screening floor, while the within-one-grade rate barely changed (95.5\% to 93.7\%).

\textbf{Conclusion:} The model preserves severity ordering under domain shift but loses grade-boundary calibration. In-domain metrics substantially overstate deployable performance, underscoring the need for external, multi-seed evaluation before clinical use.
\end{abstract}


\section{Introduction}

\subsection{Motivation}

Diabetic retinopathy is a microvascular complication of diabetes in
which sustained hyperglycaemia damages the retinal vasculature. It is
asymptomatic through its early stages, and by the time a patient notices
blurred vision, floaters, or impaired night vision, irreversible damage
is often already present [1, 2].
Diabetes affected an estimated 463 million people worldwide in 2019,
with the steepest projected growth in countries such as China and India
[3, 4, 5]. Roughly
a third of that population will develop some degree of DR [6]. Because progression is silent but treatable when caught,
periodic retinal screening is the operative intervention, and the
constraint on screening is not medicine but manpower: manual grading
requires pupil dilation, specialised lenses, and a trained
ophthalmologist, and does not scale to the population that needs it
[7].

\subsection{Problem statement}

Automated grading is therefore attractive, and deep learning has
delivered strong reported results on this task [8, 9]. But a screening algorithm is only useful if it
behaves consistently on images it has never seen, taken on cameras it
was not trained on, under illumination it has not encountered. This is
precisely where reported performance and deployed performance diverge.
Fundus images vary in illumination, colour balance, magnification, and
field of view depending on acquisition device and clinical protocol, and
models tuned on one such distribution frequently degrade on another
[10].

\subsection{Gap in existing research}

Three gaps follow from the literature reviewed in Section 2. First, most
DR grading architectures consume a single preprocessed view at a single
resolution, so they must trade global vascular context against focal
lesion detail rather than exploiting both. Second, DR severity is an
ordered scale, yet the majority of published systems optimise
categorical cross-entropy, which treats a grade-0-as-grade-4 error and a
grade-0-as-grade-1 error as equally wrong. Third, and most
consequentially, the overwhelming majority of published DR studies
report only in-domain results, so a headline accuracy figure carries no
information about screening viability and cannot be compared across
studies.

\subsection{Objective}

The objective is to construct a DR grading model that (i) extracts
complementary evidence at two resolutions through two distinct
preprocessing pipelines, (ii) respects the ordinal structure of the
severity scale, and (iii) is evaluated on a held-out multi-centre test
set with reported uncertainty, so that the magnitude of the domain-shift
penalty is measured rather than assumed away.

\subsection{Novelty and contributions}

The novelty is not any single component. Dual-branch networks, attention
fusion, and ordinal losses each exist in the literature. The novelty is
their specific combination, coupling preprocessing diversity to
resolution diversity so the two branches receive genuinely different
evidence rather than the same image at two sizes, together with an
evaluation protocol that quantifies generalization. The main
contributions are:

\textbf{(1)} A dual-resolution architecture that pairs two preprocessing
pipelines with two backbones at their native resolutions: Ben Graham
normalisation at 224x224 through EfficientNet-B0, and CLAHE at 300x300
through EfficientNet-B3 [11]. We evaluate this design
against simpler alternatives under a multi-seed protocol (Section 4.5)
and report the result plainly: adding the second backbone improves
held-out grading, whereas the attention gate and the use of two distinct
preprocessing pipelines perform within noise of simpler variants rather
than measurably better. We therefore present the fusion design as an
architectural choice, not a demonstrated accuracy gain.

\textbf{(2)} A learnable attention gate that fuses the two branches per
image, allowing the model to weight global structure against focal
detail case by case.

\textbf{(3)} An ordinal binary-decomposition head. We show the ordinal
formulation is not decorative: 95.5\% of APTOS predictions and 93.7\% of
Messidor-2 predictions fall within one grade of the reference standard,
so residual errors are clinically near-misses rather than category
collapses.

\textbf{(4)} A quantified generalization assessment. We report quadratic
weighted kappa with bootstrap confidence intervals on both an in-domain
split and a held-out multi-centre test set, show the 0.202 kappa gap
between them is statistically significant, and report referable-DR
sensitivity and specificity, the operating characteristic that
determines screening viability and that grade-wise accuracy conceals.

\section{Related Work}

\subsection{CNN-based DR grading}

Convolutional networks have been applied to DR grading extensively and
consistently outperform hand-engineered feature pipelines, because they
learn hierarchical retinal representations aligned with disease
progression [8, 9]. Mohanty et
al. [12] compared a VGG16-XGBoost hybrid against DenseNet-121 on
APTOS 2019, reporting 79.50\% and 97.30\% accuracy respectively. Raiaan
et al. [13] introduced RetNet-10 for five-class grading at 98.65\%
accuracy. Raman and Nishanthi [14] applied EfficientNet-B7 to
APTOS 2019 at 94\% accuracy, and Nivedha and Sreelakshmi [15]
reported 85.28\% with MobileNetV2. Mondal et al. [16] proposed the
EDLDR ensemble at 86.08\% for five classes and 96.98\% for the binary
task, while Nazir et al. [17] reported 97.93\% on APTOS-2019 and
98.10\% on IDRiD with a CenterNet variant, and Subbanna [18]
reported 92.6\% with a CNN-MobileNetV2 design. More recent work continues this trajectory: Shanthala and Kundur [19] proposed DR-EfficientNet-L, a multi-branch EfficientNet with attention-guided fusion that, like the present work, targets generalization across datasets; RSG-Net [20] used a dual-branch design reporting a quadratic weighted kappa of 0.930 for four-stage grading; and a 2026 systematic review [21] found that reported DR-grading accuracies vary widely across imaging sources, reinforcing the need for external validation.

The pattern across these results motivates this paper. Reported
accuracies cluster between 92\% and 99\% on the same public datasets,
yet the systems are not interchangeable in practice, and none of the
above reports performance on a held-out external cohort. The numbers are
therefore not comparable to each other, and none predicts screening
behaviour.

\subsection{Multi-scale and dual-branch architectures}

Because DR lesions span microaneurysms of a few pixels to large
haemorrhages and neovascular complexes, single-resolution models face an
unavoidable trade-off. Fu et al. [22] addressed this with
MSEF-Net, fusing features across receptive fields, and reported improved
severity grading. Shakibania et al. [23] found that a dual-branch
design with parallel feature extraction outperformed single-stream
equivalents. Both establish that parallel extraction helps; neither
pairs the branches with distinct preprocessing, so each branch sees the
same image content at a different scale rather than different evidence.

\subsection{Attention mechanisms}

Attention has been adopted in DR analysis to concentrate
representational capacity on pathological regions. Romero-Oraa et al.
[24] proposed an attention-based grading framework, and Zhang et
al. [25] applied channel and spatial attention for more selective
feature weighting. The squeeze-and-excitation mechanism that underlies
channel attention originates with Hu et al. [26]. These works
apply attention within a single feature stream; the present work instead
uses attention to arbitrate between two streams carrying different
preprocessing.

\subsection{Ordinal regression for DR grading}

DR grading is not a nominal classification problem: grades 0 through 4
are ordered, and the cost of an error scales with its distance. Niu et
al. [27] introduced the ordinal-regression CNN (OR-CNN), which
converts a K-rank problem into K-1 binary subtasks that share
intermediate layers but carry distinct weight parameters. Cao et al.
[28] later observed that this formulation permits rank
inconsistency among the binary classifiers, and proposed the COnsistent
RAnk Logits (CORAL) framework, which restores rank monotonicity by
constraining the subtasks to share a single weight vector and differ
only in their bias terms. Shi et al. [29] extended the line with
conditional probabilities. Within DR specifically, Chilukoti et al.
[30] and Karthik et al. [9] report that ordinal-aware
formulations and QWK-based evaluation better reflect clinical grading
behaviour than accuracy on nominal classes. The present work uses the
OR-CNN formulation; the implications of its weaker consistency guarantee
are addressed in Sections 3.6 and 5.1.

\subsection{Cross-dataset generalization}

Dependence on dataset-specific appearance is a recognised failure mode
in DR systems [10]. Mitigations reported in the
literature include circular cropping, contrast enhancement, and
histogram alignment [22, 24].
Voets et al. [31] attempted to reproduce a landmark DR result
using public data and documented substantial performance instability
across datasets, a finding that remains a caution to the field. The
present work takes this concern as its organising principle rather than
as a limitation paragraph.

\subsection{Positioning}

Table 1 situates this work against the studies above. The final column
is the salient one: external validation is the exception, not the norm.

\begin{longtable}[]{@{}
  >{\raggedright\arraybackslash}p{(\columnwidth - 8\tabcolsep) * \real{0.1939}}
  >{\raggedright\arraybackslash}p{(\columnwidth - 8\tabcolsep) * \real{0.2327}}
  >{\raggedright\arraybackslash}p{(\columnwidth - 8\tabcolsep) * \real{0.1828}}
  >{\raggedright\arraybackslash}p{(\columnwidth - 8\tabcolsep) * \real{0.2105}}
  >{\raggedright\arraybackslash}p{(\columnwidth - 8\tabcolsep) * \real{0.1801}}@{}}
\toprule\noalign{}
\begin{minipage}[b]{\linewidth}\raggedright
\textbf{Study}
\end{minipage} & \begin{minipage}[b]{\linewidth}\raggedright
\textbf{Method}
\end{minipage} & \begin{minipage}[b]{\linewidth}\raggedright
\textbf{Dataset}
\end{minipage} & \begin{minipage}[b]{\linewidth}\raggedright
\textbf{Reported result}
\end{minipage} & \begin{minipage}[b]{\linewidth}\raggedright
\textbf{External validation}
\end{minipage} \\
\begin{minipage}[b]{\linewidth}\raggedright
Mohanty et al. [12]
\end{minipage} & \begin{minipage}[b]{\linewidth}\raggedright
VGG16 + XGBoost / DenseNet-121
\end{minipage} & \begin{minipage}[b]{\linewidth}\raggedright
APTOS 2019
\end{minipage} & \begin{minipage}[b]{\linewidth}\raggedright
79.50\% / 97.30\% acc.
\end{minipage} & \begin{minipage}[b]{\linewidth}\raggedright
No
\end{minipage} \\
\begin{minipage}[b]{\linewidth}\raggedright
Raiaan et al. [13]
\end{minipage} & \begin{minipage}[b]{\linewidth}\raggedright
RetNet-10
\end{minipage} & \begin{minipage}[b]{\linewidth}\raggedright
Multi-source
\end{minipage} & \begin{minipage}[b]{\linewidth}\raggedright
98.65\% acc.
\end{minipage} & \begin{minipage}[b]{\linewidth}\raggedright
No
\end{minipage} \\
\begin{minipage}[b]{\linewidth}\raggedright
Nivedha \& Sreelakshmi [15]
\end{minipage} & \begin{minipage}[b]{\linewidth}\raggedright
MobileNetV2
\end{minipage} & \begin{minipage}[b]{\linewidth}\raggedright
Not stated
\end{minipage} & \begin{minipage}[b]{\linewidth}\raggedright
85.28\% acc.
\end{minipage} & \begin{minipage}[b]{\linewidth}\raggedright
No
\end{minipage} \\
\begin{minipage}[b]{\linewidth}\raggedright
Raman \& Nishanthi [14]
\end{minipage} & \begin{minipage}[b]{\linewidth}\raggedright
EfficientNet-B7
\end{minipage} & \begin{minipage}[b]{\linewidth}\raggedright
APTOS 2019
\end{minipage} & \begin{minipage}[b]{\linewidth}\raggedright
94\% acc.
\end{minipage} & \begin{minipage}[b]{\linewidth}\raggedright
No
\end{minipage} \\
\begin{minipage}[b]{\linewidth}\raggedright
Mondal et al. [16]
\end{minipage} & \begin{minipage}[b]{\linewidth}\raggedright
EDLDR ensemble
\end{minipage} & \begin{minipage}[b]{\linewidth}\raggedright
Fundus (5-class)
\end{minipage} & \begin{minipage}[b]{\linewidth}\raggedright
86.08\% acc.
\end{minipage} & \begin{minipage}[b]{\linewidth}\raggedright
No
\end{minipage} \\
\begin{minipage}[b]{\linewidth}\raggedright
Nazir et al. [17]
\end{minipage} & \begin{minipage}[b]{\linewidth}\raggedright
CenterNet
\end{minipage} & \begin{minipage}[b]{\linewidth}\raggedright
APTOS-2019, IDRiD
\end{minipage} & \begin{minipage}[b]{\linewidth}\raggedright
97.93\% / 98.10\% acc.
\end{minipage} & \begin{minipage}[b]{\linewidth}\raggedright
No
\end{minipage} \\
\begin{minipage}[b]{\linewidth}\raggedright
Subbanna [18]
\end{minipage} & \begin{minipage}[b]{\linewidth}\raggedright
CNN + MobileNetV2
\end{minipage} & \begin{minipage}[b]{\linewidth}\raggedright
Not stated
\end{minipage} & \begin{minipage}[b]{\linewidth}\raggedright
92.6\% acc.
\end{minipage} & \begin{minipage}[b]{\linewidth}\raggedright
No
\end{minipage} \\
\begin{minipage}[b]{\linewidth}\raggedright
Fu et al. [22]
\end{minipage} & \begin{minipage}[b]{\linewidth}\raggedright
MSEF-Net (multi-scale)
\end{minipage} & \begin{minipage}[b]{\linewidth}\raggedright
Fundus
\end{minipage} & \begin{minipage}[b]{\linewidth}\raggedright
Improved grading
\end{minipage} & \begin{minipage}[b]{\linewidth}\raggedright
No
\end{minipage} \\
\begin{minipage}[b]{\linewidth}\raggedright
Shakibania et al. [23]
\end{minipage} & \begin{minipage}[b]{\linewidth}\raggedright
Dual-branch CNN
\end{minipage} & \begin{minipage}[b]{\linewidth}\raggedright
Fundus
\end{minipage} & \begin{minipage}[b]{\linewidth}\raggedright
Outperforms single-stream
\end{minipage} & \begin{minipage}[b]{\linewidth}\raggedright
No
\end{minipage} \\
\begin{minipage}[b]{\linewidth}\raggedright
\textbf{This work}
\end{minipage} & \begin{minipage}[b]{\linewidth}\raggedright
\textbf{Dual-resolution + attention gate + ordinal head}
\end{minipage} & \begin{minipage}[b]{\linewidth}\raggedright
\textbf{APTOS 2019 + Messidor-2}
\end{minipage} & \begin{minipage}[b]{\linewidth}\raggedright
\textbf{QWK 0.882 / 0.679 with 95\% CIs}
\end{minipage} & \begin{minipage}[b]{\linewidth}\raggedright
\textbf{Yes (held-out)}
\end{minipage} \\
\midrule\noalign{}
\endhead
\bottomrule\noalign{}
\endlastfoot
\end{longtable}

\textbf{Table 1} Positioning relative to recent DR grading literature.
Accuracies are as reported by the original authors on their own
protocols and are not directly comparable across rows.

\section{Materials and Methods}

\subsection{Overview}

The pipeline comprises six stages: dataset preparation and splitting
(3.2); preprocessing and domain normalisation (3.3); dual-resolution
feature extraction (3.4); attention-gated fusion (3.5); ordinal
regression (3.6); and training and evaluation (3.7-3.9). The design
intent throughout is that the two branches receive complementary rather
than redundant evidence, and that no test-set information influences any
modelling decision.

\subsection{Datasets and splits}

Two public fundus collections were used. APTOS 2019 [32]
contributed 3,662 images graded 0 (no DR) to 4 (proliferative DR), split
80/20 by stratified sampling (random seed 42) into 2,929 training and
733 validation images. Messidor-2 [33]
comprises images from multiple clinical centres acquired on different
devices, with correspondingly wider variation in illumination, colour
distribution, and field of view. Retaining only images flagged as
adjudicated-gradable and resolvable on disk left 1,744 images, split
70/30 by stratified sampling (seed 42) into 1,220 training and 524 test
images. The training set combined the APTOS training portion with the
Messidor-2 training portion, giving 4,149 images. The Messidor-2 test
portion was excluded from training and from every model-selection
decision. Table 2 reports the composition.

\begin{longtable}[]{@{}
  >{\raggedright\arraybackslash}p{(\columnwidth - 16\tabcolsep) * \real{0.2023}}
  >{\raggedright\arraybackslash}p{(\columnwidth - 16\tabcolsep) * \real{0.1850}}
  >{\raggedright\arraybackslash}p{(\columnwidth - 16\tabcolsep) * \real{0.1734}}
  >{\raggedright\arraybackslash}p{(\columnwidth - 16\tabcolsep) * \real{0.0809}}
  >{\raggedright\arraybackslash}p{(\columnwidth - 16\tabcolsep) * \real{0.0717}}
  >{\raggedright\arraybackslash}p{(\columnwidth - 16\tabcolsep) * \real{0.0717}}
  >{\raggedright\arraybackslash}p{(\columnwidth - 16\tabcolsep) * \real{0.0717}}
  >{\raggedright\arraybackslash}p{(\columnwidth - 16\tabcolsep) * \real{0.0717}}
  >{\raggedright\arraybackslash}p{(\columnwidth - 16\tabcolsep) * \real{0.0717}}@{}}
\toprule\noalign{}
\begin{minipage}[b]{\linewidth}\raggedright
\textbf{Partition}
\end{minipage} & \begin{minipage}[b]{\linewidth}\raggedright
\textbf{Source}
\end{minipage} & \begin{minipage}[b]{\linewidth}\raggedright
\textbf{Role}
\end{minipage} & \begin{minipage}[b]{\linewidth}\raggedright
\textbf{n}
\end{minipage} & \begin{minipage}[b]{\linewidth}\raggedright
\textbf{G0}
\end{minipage} & \begin{minipage}[b]{\linewidth}\raggedright
\textbf{G1}
\end{minipage} & \begin{minipage}[b]{\linewidth}\raggedright
\textbf{G2}
\end{minipage} & \begin{minipage}[b]{\linewidth}\raggedright
\textbf{G3}
\end{minipage} & \begin{minipage}[b]{\linewidth}\raggedright
\textbf{G4}
\end{minipage} \\
\begin{minipage}[b]{\linewidth}\raggedright
APTOS 2019 (train)
\end{minipage} & \begin{minipage}[b]{\linewidth}\raggedright
Single centre, India
\end{minipage} & \begin{minipage}[b]{\linewidth}\raggedright
Training
\end{minipage} & \begin{minipage}[b]{\linewidth}\raggedright
2,929
\end{minipage} & \begin{minipage}[b]{\linewidth}\raggedright
1,444
\end{minipage} & \begin{minipage}[b]{\linewidth}\raggedright
296
\end{minipage} & \begin{minipage}[b]{\linewidth}\raggedright
799
\end{minipage} & \begin{minipage}[b]{\linewidth}\raggedright
154
\end{minipage} & \begin{minipage}[b]{\linewidth}\raggedright
236
\end{minipage} \\
\begin{minipage}[b]{\linewidth}\raggedright
APTOS 2019 (val)
\end{minipage} & \begin{minipage}[b]{\linewidth}\raggedright
Single centre, India
\end{minipage} & \begin{minipage}[b]{\linewidth}\raggedright
Model selection
\end{minipage} & \begin{minipage}[b]{\linewidth}\raggedright
733
\end{minipage} & \begin{minipage}[b]{\linewidth}\raggedright
361
\end{minipage} & \begin{minipage}[b]{\linewidth}\raggedright
74
\end{minipage} & \begin{minipage}[b]{\linewidth}\raggedright
200
\end{minipage} & \begin{minipage}[b]{\linewidth}\raggedright
39
\end{minipage} & \begin{minipage}[b]{\linewidth}\raggedright
59
\end{minipage} \\
\begin{minipage}[b]{\linewidth}\raggedright
Messidor-2 (train)
\end{minipage} & \begin{minipage}[b]{\linewidth}\raggedright
Multi-centre, France
\end{minipage} & \begin{minipage}[b]{\linewidth}\raggedright
Training
\end{minipage} & \begin{minipage}[b]{\linewidth}\raggedright
1,220
\end{minipage} & \begin{minipage}[b]{\linewidth}\raggedright
711
\end{minipage} & \begin{minipage}[b]{\linewidth}\raggedright
189
\end{minipage} & \begin{minipage}[b]{\linewidth}\raggedright
243
\end{minipage} & \begin{minipage}[b]{\linewidth}\raggedright
52
\end{minipage} & \begin{minipage}[b]{\linewidth}\raggedright
25
\end{minipage} \\
\begin{minipage}[b]{\linewidth}\raggedright
\textbf{Messidor-2 (test)}
\end{minipage} & \begin{minipage}[b]{\linewidth}\raggedright
\textbf{Multi-centre, France}
\end{minipage} & \begin{minipage}[b]{\linewidth}\raggedright
\textbf{Held-out eval.}
\end{minipage} & \begin{minipage}[b]{\linewidth}\raggedright
\textbf{524}
\end{minipage} & \begin{minipage}[b]{\linewidth}\raggedright
\textbf{306}
\end{minipage} & \begin{minipage}[b]{\linewidth}\raggedright
\textbf{81}
\end{minipage} & \begin{minipage}[b]{\linewidth}\raggedright
\textbf{104}
\end{minipage} & \begin{minipage}[b]{\linewidth}\raggedright
\textbf{23}
\end{minipage} & \begin{minipage}[b]{\linewidth}\raggedright
\textbf{10}
\end{minipage} \\
\midrule\noalign{}
\endhead
\bottomrule\noalign{}
\endlastfoot
\end{longtable}

\textbf{Table 2} Dataset composition and partitioning. G0-G4 denote DR severity
grades 0-4. Combined training set n = 4,149. All counts are exact. Figure 1 shows representative fundus images for each severity grade.

\includegraphics[width=\textwidth]{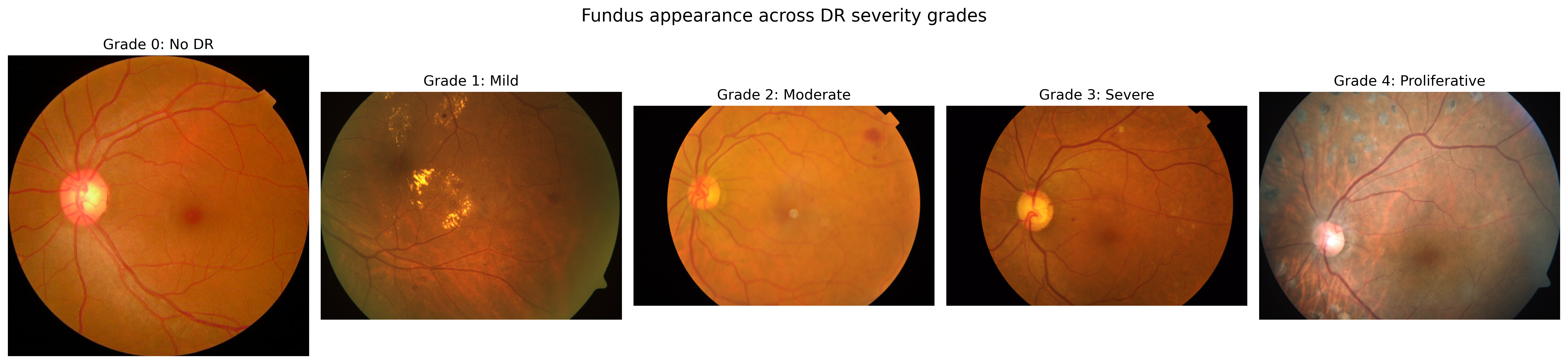}

\textbf{Fig. 1} Representative fundus images across the five DR severity
grades, from no DR (grade 0) to proliferative DR (grade 4)

\subsection{Preprocessing and domain normalisation}

\subsubsection{Retinal region isolation}

Each image was cropped circularly to the retinal disc, removing the
uninformative dark border and standardising spatial layout across
acquisition devices with differing sensor aspect ratios.

\subsubsection{Dataset-aware enhancement}

Two enhancement pipelines produce the two branch inputs, chosen because
they emphasise different pathology. Ben Graham normalisation [34] subtracts a local Gaussian-blurred estimate to suppress
illumination gradients and accentuate vascular structure.
Contrast-Limited Adaptive Histogram Equalisation (CLAHE) [35] amplifies local contrast within tiles, improving the visibility
of microaneurysms and small exudates that global normalisation tends to
flatten. The pairing is deliberate: the Ben Graham view carries
structure, the CLAHE view carries focal lesions, and Section 3.5 lets
the model decide which to trust per image.

\subsubsection{Histogram matching for domain alignment}

To reduce colour and intensity discrepancy between collections,
histogram matching [36] was applied to
Messidor-2 images only, against a single reference image drawn from the
APTOS training split, aligning first-order intensity statistics without
altering anatomical content. The reference was the first image of the
shuffled APTOS training partition, selected arbitrarily rather than by
any image-quality criterion; the same reference was applied to
Messidor-2 training and test images alike, so no test-set statistics
enter the transform. APTOS images were not histogram-matched.

\subsubsection{Augmentation}

Applied to training images only: random horizontal flipping (p = 0.5),
random brightness and contrast jitter (p = 0.4, limits ±0.2 on each),
and random gamma correction (p = 0.3, gamma range 0.8-1.2). Both branch
views of an image are augmented independently. Validation and test
images received resizing and normalisation only. Figure 2 illustrates the preprocessing pipeline on a representative image.

\includegraphics[width=\textwidth]{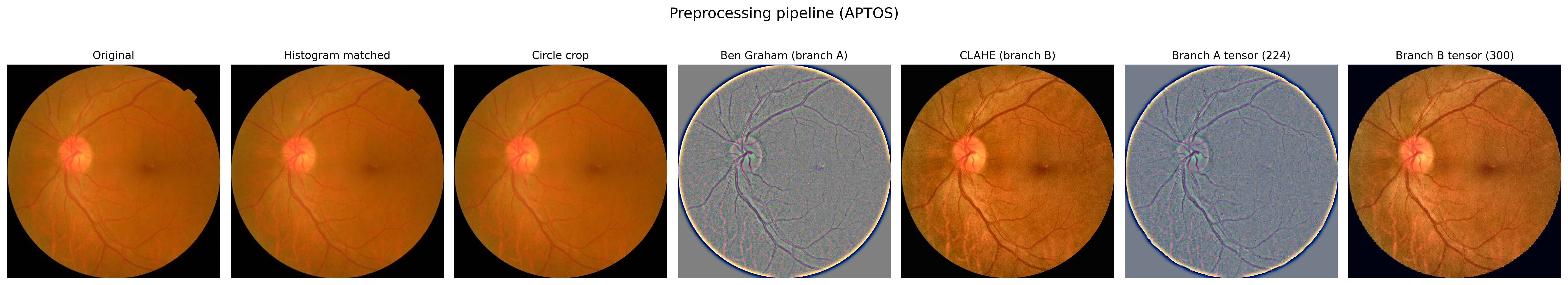}

\textbf{Fig. 2} The preprocessing pipeline on a single APTOS image, from
the original through circle-cropping to the two branch inputs: Ben
Graham normalisation (branch A, 224x224) and CLAHE (branch B, 300x300),
with the normalised tensors each branch receives. Figure 3 shows this pipeline applied to images from both source domains.

\includegraphics[width=\textwidth]{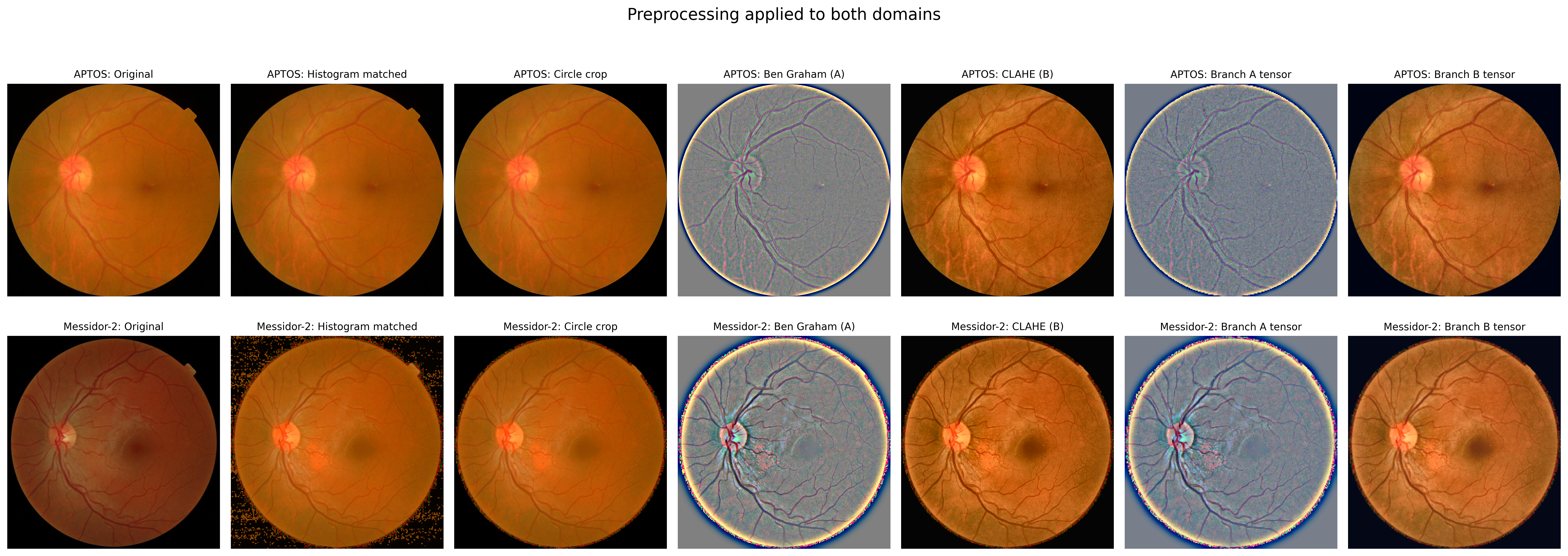}

\textbf{Fig. 3} The same pipeline applied to both domains, APTOS (top)
and Messidor-2 (bottom). Messidor-2 additionally undergoes histogram
matching to an APTOS reference. The residual differences in colour and
illumination after normalisation illustrate the domain shift the model
must bridge

\subsection{Dual-resolution feature extraction}

Two ImageNet-pretrained EfficientNet backbones [11]
run in parallel, instantiated with their classifier heads removed so
that each emits a pooled feature vector. The B0 branch consumes the Ben
Graham view at 224x224 and yields f₀ ∈ R\^{}1280; the B3 branch consumes
the CLAHE view at 300x300 and yields f₃ ∈ R\^{}1536. Both backbones are
fine-tuned end to end; no layers are frozen.

The resolutions are not arbitrary: 224x224 and 300x300 are the native
training resolutions of EfficientNet-B0 and B3 respectively under the
compound-scaling rule of Tan and Le [11]. Each backbone therefore
operates at the resolution its depth and width were jointly scaled for.
Pairing the lower-resolution backbone with the structure-emphasising
view and the higher-resolution backbone with the lesion-emphasising view
follows from the same logic: fine lesion detail is the component that
requires spatial resolution to survive.

\subsection{Attention-gated fusion}

The two feature vectors are concatenated and passed through a two-layer
bottleneck in the style of squeeze-and-excitation [26]
to produce a per-channel gating vector:

\emph{f\_cat = {[}f₀ ; f₃{]} ∈ R\^{}2816 , g = σ(W₂ δ(W₁ f\_cat)) ∈
R\^{}1280} (1)

where W₁ ∈ R\^{}(256 x 2816), W₂ ∈ R\^{}(1280 x 256), δ is ReLU and σ is
the logistic sigmoid. The gate is therefore computed from both branches
jointly but is dimensioned to the B0 branch. Fusion is a per-channel
convex combination:

\emph{f\_fused = f₀ ⊙ g + f₃\^{}(1:1280) ⊙ (1 − g) ∈ R\^{}1280} (2)

where f₃\^{}(1:1280) denotes the first 1,280 components of f₃. Because
the two backbones emit vectors of different width, the wider B3 vector
is truncated to the width of the B0 vector so that the elementwise
combination is defined. This is a deliberate simplification with two
consequences that should be stated plainly: 256 of the 1,536 B3 channels
(16.7\%) are computed and then discarded, and the channel indices of the
two backbones are placed in correspondence without any learned alignment
between them. A learned projection of both vectors into a shared space
would remove both objections; Section 5.5 records this as a limitation,
and Tables 8 and 9 evaluate the fusion component against a
plain-concatenation alternative.

Because g is produced from both branches jointly and applied per
channel, the model can rely on global structure for one image and focal
detail for another, and can mix them channel-wise within a single image.
The convex form keeps the fused vector bounded by the branch features
rather than growing in magnitude with branch agreement.

\subsection{Ordinal regression head}

DR severity is an ordered scale, so the head is an ordinal binary
decomposition rather than a softmax classifier. The fused vector passes
through a bottleneck, a linear map to 128 units, ReLU, and dropout
with rate 0.3, whose output h ∈ R\^{}128 feeds K-1 = 4 binary
decision functions, one per threshold between successive grades:

\emph{z\_k = w\_kᵀ h + b\_k , P(y \textgreater{} r\_k \textbar{} x) =
σ(z\_k) , k = 1 ... K−1} (3)

Training minimises the mean binary cross-entropy over the four subtasks,
unweighted (λ\_k = 1 for all k):

\emph{L = − (1/4) Σₙ Σₖ {[} yₖⁿ log σ(zₖⁿ) + (1 − yₖⁿ) log(1 − σ(zₖⁿ))
{]} , yₖⁿ = 1\{yⁿ \textgreater{} rₖ\}} (4)

At inference the grade is recovered by counting satisfied thresholds:

\emph{ŷ = Σₖ 1\{σ(zₖ) \textgreater{} 0.5\}} (5)

This is the OR-CNN formulation of Niu et al. [27]: the four
subtasks share all intermediate layers but carry distinct weight vectors
w\_k. It is important to be precise about what this does and does not
guarantee. Because the w\_k are learned independently, the predicted
exceedance probabilities are not constrained to be monotonically
decreasing in k, so rank-inconsistent output patterns are representable,
and the threshold-counting rule of Equation 5 aggregates them without
detecting the inconsistency. The rank-monotonicity guarantee of CORAL
[28] requires the stronger constraint w\_k = w for all
k, which this implementation does not impose. The high within-one-grade
rates reported in Section 4 are therefore an empirical property of the
trained model, not a structural consequence of the head, and are
described as such throughout.

\subsection{Implementation details}

Table 3 lists the configuration. Class imbalance was handled by a
weighted random sampler drawing with replacement at inverse label
frequency over the combined training set, rather than by reweighting the
loss.

\begin{longtable}[]{@{}
  >{\raggedright\arraybackslash}p{(\columnwidth - 2\tabcolsep) * \real{0.3545}}
  >{\raggedright\arraybackslash}p{(\columnwidth - 2\tabcolsep) * \real{0.6455}}@{}}
\toprule\noalign{}
\begin{minipage}[b]{\linewidth}\raggedright
\textbf{Component}
\end{minipage} & \begin{minipage}[b]{\linewidth}\raggedright
\textbf{Setting}
\end{minipage} \\
\begin{minipage}[b]{\linewidth}\raggedright
Branch A backbone
\end{minipage} & \begin{minipage}[b]{\linewidth}\raggedright
EfficientNet-B0 (timm), ImageNet-pretrained, head removed
\end{minipage} \\
\begin{minipage}[b]{\linewidth}\raggedright
Branch B backbone
\end{minipage} & \begin{minipage}[b]{\linewidth}\raggedright
EfficientNet-B3 (timm), ImageNet-pretrained, head removed
\end{minipage} \\
\begin{minipage}[b]{\linewidth}\raggedright
Branch A input
\end{minipage} & \begin{minipage}[b]{\linewidth}\raggedright
Ben Graham normalised, 224 x 224
\end{minipage} \\
\begin{minipage}[b]{\linewidth}\raggedright
Branch B input
\end{minipage} & \begin{minipage}[b]{\linewidth}\raggedright
CLAHE enhanced, 300 x 300
\end{minipage} \\
\begin{minipage}[b]{\linewidth}\raggedright
Feature widths
\end{minipage} & \begin{minipage}[b]{\linewidth}\raggedright
1,280 (B0) and 1,536 (B3)
\end{minipage} \\
\begin{minipage}[b]{\linewidth}\raggedright
Attention bottleneck
\end{minipage} & \begin{minipage}[b]{\linewidth}\raggedright
Linear 2816 → 256, ReLU, Linear 256 → 1280, sigmoid
\end{minipage} \\
\begin{minipage}[b]{\linewidth}\raggedright
Fusion
\end{minipage} & \begin{minipage}[b]{\linewidth}\raggedright
Per-channel convex combination; B3 truncated to 1,280
\end{minipage} \\
\begin{minipage}[b]{\linewidth}\raggedright
Classifier bottleneck
\end{minipage} & \begin{minipage}[b]{\linewidth}\raggedright
Linear 1280 → 128, ReLU, dropout 0.3
\end{minipage} \\
\begin{minipage}[b]{\linewidth}\raggedright
Ordinal head
\end{minipage} & \begin{minipage}[b]{\linewidth}\raggedright
4 independent binary logits (OR-CNN)
\end{minipage} \\
\begin{minipage}[b]{\linewidth}\raggedright
Loss
\end{minipage} & \begin{minipage}[b]{\linewidth}\raggedright
Mean binary cross-entropy with logits, unweighted
\end{minipage} \\
\begin{minipage}[b]{\linewidth}\raggedright
Optimiser
\end{minipage} & \begin{minipage}[b]{\linewidth}\raggedright
Adam
\end{minipage} \\
\begin{minipage}[b]{\linewidth}\raggedright
Learning rate
\end{minipage} & \begin{minipage}[b]{\linewidth}\raggedright
1 x 10⁻⁴, constant (no schedule)
\end{minipage} \\
\begin{minipage}[b]{\linewidth}\raggedright
Weight decay
\end{minipage} & \begin{minipage}[b]{\linewidth}\raggedright
None (0)
\end{minipage} \\
\begin{minipage}[b]{\linewidth}\raggedright
Batch size
\end{minipage} & \begin{minipage}[b]{\linewidth}\raggedright
16
\end{minipage} \\
\begin{minipage}[b]{\linewidth}\raggedright
Epochs
\end{minipage} & \begin{minipage}[b]{\linewidth}\raggedright
10; best-validation-QWK checkpoint (epoch 8) retained
\end{minipage} \\
\begin{minipage}[b]{\linewidth}\raggedright
Precision
\end{minipage} & \begin{minipage}[b]{\linewidth}\raggedright
Mixed precision (AMP) with gradient scaling
\end{minipage} \\
\begin{minipage}[b]{\linewidth}\raggedright
Class balancing
\end{minipage} & \begin{minipage}[b]{\linewidth}\raggedright
Weighted random sampler, inverse label frequency, with replacement
\end{minipage} \\
\begin{minipage}[b]{\linewidth}\raggedright
Backbone freezing
\end{minipage} & \begin{minipage}[b]{\linewidth}\raggedright
None; end-to-end fine-tuning
\end{minipage} \\
\begin{minipage}[b]{\linewidth}\raggedright
Framework
\end{minipage} & \begin{minipage}[b]{\linewidth}\raggedright
PyTorch with timm and albumentations
\end{minipage} \\
\begin{minipage}[b]{\linewidth}\raggedright
Hardware
\end{minipage} & \begin{minipage}[b]{\linewidth}\raggedright
Single NVIDIA GPU (Kaggle runtime)
\end{minipage} \\
\begin{minipage}[b]{\linewidth}\raggedright
Training wall-clock
\end{minipage} & \begin{minipage}[b]{\linewidth}\raggedright
≈ 24.5 min per epoch; ≈ 4.1 h total
\end{minipage} \\
\begin{minipage}[b]{\linewidth}\raggedright
Random seed
\end{minipage} & \begin{minipage}[b]{\linewidth}\raggedright
42 (dataset splits)
\end{minipage} \\
\midrule\noalign{}
\endhead
\bottomrule\noalign{}
\endlastfoot
\end{longtable}

\textbf{Table 3} Implementation configuration.

\subsection{Hyperparameter selection strategy}

Model parameters were selected in three tiers. First, values inherited
from source publications and not tuned: the 224x224 and 300x300 input
resolutions are the native resolutions of EfficientNet-B0 and B3 under
compound scaling [11], and the ordinal binary
decomposition is taken from Niu et al. [27]. Second, values set to
community-standard defaults and adopted without search: Adam at a
constant learning rate of 1 x 10⁻⁴ with no weight decay, batch size 16,
dropout 0.3, and the 256- and 128-unit bottleneck widths. Third, the
single quantity selected empirically: the training epoch, chosen by
monitoring quadratic weighted kappa on the APTOS validation split and
retaining the best-scoring checkpoint. No hyperparameter sweep was
performed. The Messidor-2 test set played no role in any tier: no
architecture, hyperparameter, threshold, or checkpoint was chosen with
reference to it.

\subsection{Evaluation metrics}

Quadratic weighted kappa is the primary metric, as it penalises errors
in proportion to the squared distance between predicted and reference
grades and thus reflects the ordinal cost structure of clinical grading.
We additionally report overall accuracy, linear kappa, the proportion of
predictions within one grade of reference, and per-grade precision,
recall, and F1. Referable DR (grade \textgreater= 2) sensitivity and
specificity are reported separately, as this binary operating point
determines whether a system is usable for screening triage. Inference is
a single deterministic forward pass per image at the operating threshold
of Equation 5; no test-time augmentation, ensembling, or post-hoc grade
adjustment is applied. Uncertainty is quantified by bootstrap resampling
of the evaluation sets with 5,000 replicates, reported as 95\%
percentile intervals.

\section{Results}

\subsection{Training dynamics}

Table 4 reports the full training trajectory. Training loss fell
monotonically from 0.371 to 0.045 and training QWK rose from 0.719 to
0.982, while validation QWK rose quickly to 0.871 by epoch 4 and then
plateaued, peaking at 0.8815 at epoch 8 and falling back to 0.864 over
the final two epochs. Two observations follow. First, the checkpoint
selection is not an artefact of stopping at an arbitrary epoch:
validation performance had flattened by epoch 4 and declined after epoch
8, so additional epochs under this configuration would not have improved
the model. Second, the terminal gap between training QWK (0.982) and
validation QWK (0.882) indicates the configuration is capacity-limited
rather than convergence-limited, the model has enough capacity to fit
the training distribution closely without transferring that fit to
held-out data.

\begin{longtable}[]{@{}
  >{\raggedright\arraybackslash}p{(\columnwidth - 6\tabcolsep) * \real{0.1329}}
  >{\raggedright\arraybackslash}p{(\columnwidth - 6\tabcolsep) * \real{0.2881}}
  >{\raggedright\arraybackslash}p{(\columnwidth - 6\tabcolsep) * \real{0.2881}}
  >{\raggedright\arraybackslash}p{(\columnwidth - 6\tabcolsep) * \real{0.2909}}@{}}
\toprule\noalign{}
\begin{minipage}[b]{\linewidth}\raggedright
\textbf{Epoch}
\end{minipage} & \begin{minipage}[b]{\linewidth}\raggedright
\textbf{Training loss}
\end{minipage} & \begin{minipage}[b]{\linewidth}\raggedright
\textbf{Training QWK}
\end{minipage} & \begin{minipage}[b]{\linewidth}\raggedright
\textbf{APTOS validation QWK}
\end{minipage} \\
\begin{minipage}[b]{\linewidth}\raggedright
1
\end{minipage} & \begin{minipage}[b]{\linewidth}\raggedright
0.371
\end{minipage} & \begin{minipage}[b]{\linewidth}\raggedright
0.719
\end{minipage} & \begin{minipage}[b]{\linewidth}\raggedright
0.864
\end{minipage} \\
\begin{minipage}[b]{\linewidth}\raggedright
2
\end{minipage} & \begin{minipage}[b]{\linewidth}\raggedright
0.193
\end{minipage} & \begin{minipage}[b]{\linewidth}\raggedright
0.903
\end{minipage} & \begin{minipage}[b]{\linewidth}\raggedright
0.852
\end{minipage} \\
\begin{minipage}[b]{\linewidth}\raggedright
3
\end{minipage} & \begin{minipage}[b]{\linewidth}\raggedright
0.130
\end{minipage} & \begin{minipage}[b]{\linewidth}\raggedright
0.936
\end{minipage} & \begin{minipage}[b]{\linewidth}\raggedright
0.863
\end{minipage} \\
\begin{minipage}[b]{\linewidth}\raggedright
4
\end{minipage} & \begin{minipage}[b]{\linewidth}\raggedright
0.113
\end{minipage} & \begin{minipage}[b]{\linewidth}\raggedright
0.946
\end{minipage} & \begin{minipage}[b]{\linewidth}\raggedright
0.871
\end{minipage} \\
\begin{minipage}[b]{\linewidth}\raggedright
5
\end{minipage} & \begin{minipage}[b]{\linewidth}\raggedright
0.077
\end{minipage} & \begin{minipage}[b]{\linewidth}\raggedright
0.969
\end{minipage} & \begin{minipage}[b]{\linewidth}\raggedright
0.875
\end{minipage} \\
\begin{minipage}[b]{\linewidth}\raggedright
6
\end{minipage} & \begin{minipage}[b]{\linewidth}\raggedright
0.069
\end{minipage} & \begin{minipage}[b]{\linewidth}\raggedright
0.970
\end{minipage} & \begin{minipage}[b]{\linewidth}\raggedright
0.872
\end{minipage} \\
\begin{minipage}[b]{\linewidth}\raggedright
7
\end{minipage} & \begin{minipage}[b]{\linewidth}\raggedright
0.053
\end{minipage} & \begin{minipage}[b]{\linewidth}\raggedright
0.978
\end{minipage} & \begin{minipage}[b]{\linewidth}\raggedright
0.881
\end{minipage} \\
\begin{minipage}[b]{\linewidth}\raggedright
\textbf{8}
\end{minipage} & \begin{minipage}[b]{\linewidth}\raggedright
\textbf{0.045}
\end{minipage} & \begin{minipage}[b]{\linewidth}\raggedright
\textbf{0.982}
\end{minipage} & \begin{minipage}[b]{\linewidth}\raggedright
\subsection{(best; retained)}
\end{minipage} \\
\begin{minipage}[b]{\linewidth}\raggedright
9
\end{minipage} & \begin{minipage}[b]{\linewidth}\raggedright
0.047
\end{minipage} & \begin{minipage}[b]{\linewidth}\raggedright
0.979
\end{minipage} & \begin{minipage}[b]{\linewidth}\raggedright
0.864
\end{minipage} \\
\begin{minipage}[b]{\linewidth}\raggedright
10
\end{minipage} & \begin{minipage}[b]{\linewidth}\raggedright
0.048
\end{minipage} & \begin{minipage}[b]{\linewidth}\raggedright
0.981
\end{minipage} & \begin{minipage}[b]{\linewidth}\raggedright
0.864
\end{minipage} \\
\midrule\noalign{}
\endhead
\bottomrule\noalign{}
\endlastfoot
\end{longtable}

\textbf{Table 4} Training trajectory over 10 epochs. The epoch-8 checkpoint was
retained for all subsequent evaluation. Figure 4 shows the training trajectory.

\includegraphics[width=\textwidth]{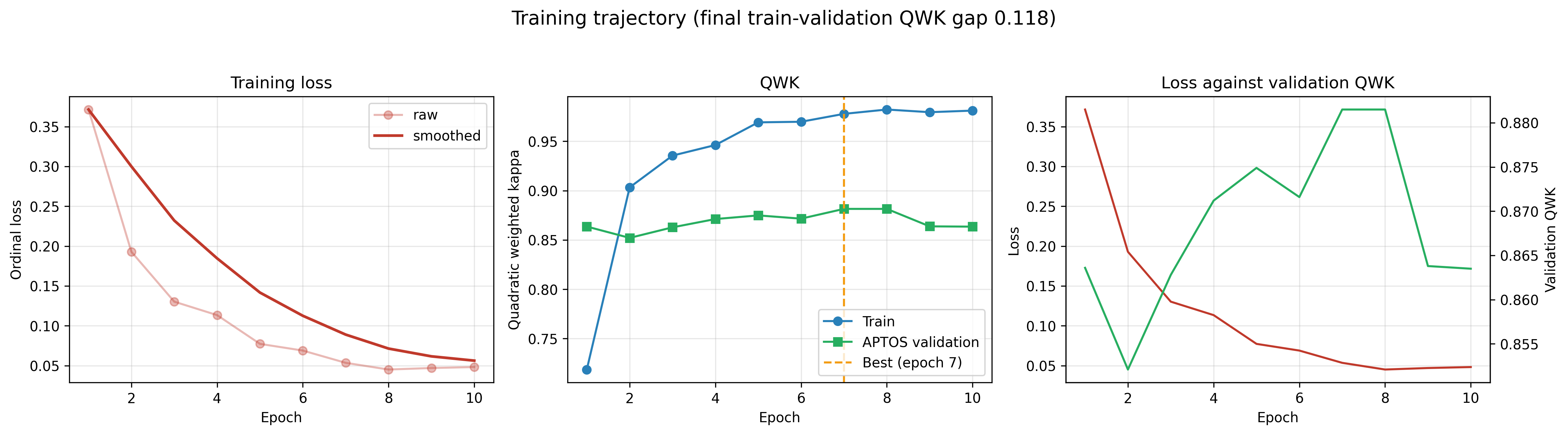}

\textbf{Fig. 4} Training trajectory of the full model. Left: ordinal
loss per epoch. Centre: quadratic weighted kappa on the training and
APTOS validation splits, with the retained epoch marked. Right: loss
against validation kappa. Validation kappa plateaus after epoch 4 and
the best checkpoint is at epoch 8

\subsection{In-domain performance on APTOS}

On the held-out APTOS validation split (n = 733) the retained model
achieved QWK 0.882 (95\% CI 0.853-0.906) and accuracy 80.6\% (95\% CI
77.8-83.5\%). Per-grade results are given in Table 5. Errors concentrate
overwhelmingly between adjacent grades: 95.5\% of predictions fall
within one grade of the reference standard. Performance is strongly
grade-dependent. Grade 0 is near-ceiling (F1 0.963), while grades 1 and
3, the least represented and the most clinically ambiguous, are
markedly weaker (F1 0.542 and 0.380).

\begin{longtable}[]{@{}
  >{\raggedright\arraybackslash}p{(\columnwidth - 10\tabcolsep) * \real{0.0997}}
  >{\raggedright\arraybackslash}p{(\columnwidth - 10\tabcolsep) * \real{0.2659}}
  >{\raggedright\arraybackslash}p{(\columnwidth - 10\tabcolsep) * \real{0.0997}}
  >{\raggedright\arraybackslash}p{(\columnwidth - 10\tabcolsep) * \real{0.1773}}
  >{\raggedright\arraybackslash}p{(\columnwidth - 10\tabcolsep) * \real{0.1773}}
  >{\raggedright\arraybackslash}p{(\columnwidth - 10\tabcolsep) * \real{0.1801}}@{}}
\toprule\noalign{}
\begin{minipage}[b]{\linewidth}\raggedright
\textbf{Grade}
\end{minipage} & \begin{minipage}[b]{\linewidth}\raggedright
\textbf{Description}
\end{minipage} & \begin{minipage}[b]{\linewidth}\raggedright
\textbf{n}
\end{minipage} & \begin{minipage}[b]{\linewidth}\raggedright
\textbf{Precision}
\end{minipage} & \begin{minipage}[b]{\linewidth}\raggedright
\textbf{Recall}
\end{minipage} & \begin{minipage}[b]{\linewidth}\raggedright
\textbf{F1}
\end{minipage} \\
\begin{minipage}[b]{\linewidth}\raggedright
0
\end{minipage} & \begin{minipage}[b]{\linewidth}\raggedright
No DR
\end{minipage} & \begin{minipage}[b]{\linewidth}\raggedright
361
\end{minipage} & \begin{minipage}[b]{\linewidth}\raggedright
0.961
\end{minipage} & \begin{minipage}[b]{\linewidth}\raggedright
0.964
\end{minipage} & \begin{minipage}[b]{\linewidth}\raggedright
0.963
\end{minipage} \\
\begin{minipage}[b]{\linewidth}\raggedright
1
\end{minipage} & \begin{minipage}[b]{\linewidth}\raggedright
Mild NPDR
\end{minipage} & \begin{minipage}[b]{\linewidth}\raggedright
74
\end{minipage} & \begin{minipage}[b]{\linewidth}\raggedright
0.489
\end{minipage} & \begin{minipage}[b]{\linewidth}\raggedright
0.608
\end{minipage} & \begin{minipage}[b]{\linewidth}\raggedright
0.542
\end{minipage} \\
\begin{minipage}[b]{\linewidth}\raggedright
2
\end{minipage} & \begin{minipage}[b]{\linewidth}\raggedright
Moderate NPDR
\end{minipage} & \begin{minipage}[b]{\linewidth}\raggedright
200
\end{minipage} & \begin{minipage}[b]{\linewidth}\raggedright
0.786
\end{minipage} & \begin{minipage}[b]{\linewidth}\raggedright
0.755
\end{minipage} & \begin{minipage}[b]{\linewidth}\raggedright
0.770
\end{minipage} \\
\begin{minipage}[b]{\linewidth}\raggedright
3
\end{minipage} & \begin{minipage}[b]{\linewidth}\raggedright
Severe NPDR
\end{minipage} & \begin{minipage}[b]{\linewidth}\raggedright
39
\end{minipage} & \begin{minipage}[b]{\linewidth}\raggedright
0.375
\end{minipage} & \begin{minipage}[b]{\linewidth}\raggedright
0.385
\end{minipage} & \begin{minipage}[b]{\linewidth}\raggedright
0.380
\end{minipage} \\
\begin{minipage}[b]{\linewidth}\raggedright
4
\end{minipage} & \begin{minipage}[b]{\linewidth}\raggedright
Proliferative DR
\end{minipage} & \begin{minipage}[b]{\linewidth}\raggedright
59
\end{minipage} & \begin{minipage}[b]{\linewidth}\raggedright
0.681
\end{minipage} & \begin{minipage}[b]{\linewidth}\raggedright
0.542
\end{minipage} & \begin{minipage}[b]{\linewidth}\raggedright
0.604
\end{minipage} \\
\midrule\noalign{}
\endhead
\bottomrule\noalign{}
\endlastfoot
\end{longtable}

\textbf{Table 5} Per-grade performance on the APTOS validation split (n = 733).

\subsection{Held-out performance on Messidor-2}

On the Messidor-2 test set (n = 524), excluded from training and from
all model selection, the model achieved QWK 0.679 (95\% CI 0.613-0.735)
and accuracy 61.3\% (95\% CI 57.1-65.5\%). Per-grade results appear in
\textbf{Table 6} The degradation is not uniform. Grade 0 remains serviceable (F1
0.759), but grade 1 collapses to F1 0.264, with 71 of 306 true grade-0
images misassigned to grade 1 and only 28 of 81 true grade-1 images
recovered. Grade 4 recall falls to 0.300 on a support of 10 images.

\begin{longtable}[]{@{}
  >{\raggedright\arraybackslash}p{(\columnwidth - 10\tabcolsep) * \real{0.0997}}
  >{\raggedright\arraybackslash}p{(\columnwidth - 10\tabcolsep) * \real{0.2659}}
  >{\raggedright\arraybackslash}p{(\columnwidth - 10\tabcolsep) * \real{0.0997}}
  >{\raggedright\arraybackslash}p{(\columnwidth - 10\tabcolsep) * \real{0.1773}}
  >{\raggedright\arraybackslash}p{(\columnwidth - 10\tabcolsep) * \real{0.1773}}
  >{\raggedright\arraybackslash}p{(\columnwidth - 10\tabcolsep) * \real{0.1801}}@{}}
\toprule\noalign{}
\begin{minipage}[b]{\linewidth}\raggedright
\textbf{Grade}
\end{minipage} & \begin{minipage}[b]{\linewidth}\raggedright
\textbf{Description}
\end{minipage} & \begin{minipage}[b]{\linewidth}\raggedright
\textbf{n}
\end{minipage} & \begin{minipage}[b]{\linewidth}\raggedright
\textbf{Precision}
\end{minipage} & \begin{minipage}[b]{\linewidth}\raggedright
\textbf{Recall}
\end{minipage} & \begin{minipage}[b]{\linewidth}\raggedright
\textbf{F1}
\end{minipage} \\
\begin{minipage}[b]{\linewidth}\raggedright
0
\end{minipage} & \begin{minipage}[b]{\linewidth}\raggedright
No DR
\end{minipage} & \begin{minipage}[b]{\linewidth}\raggedright
306
\end{minipage} & \begin{minipage}[b]{\linewidth}\raggedright
0.773
\end{minipage} & \begin{minipage}[b]{\linewidth}\raggedright
0.745
\end{minipage} & \begin{minipage}[b]{\linewidth}\raggedright
0.759
\end{minipage} \\
\begin{minipage}[b]{\linewidth}\raggedright
1
\end{minipage} & \begin{minipage}[b]{\linewidth}\raggedright
Mild NPDR
\end{minipage} & \begin{minipage}[b]{\linewidth}\raggedright
81
\end{minipage} & \begin{minipage}[b]{\linewidth}\raggedright
0.214
\end{minipage} & \begin{minipage}[b]{\linewidth}\raggedright
0.346
\end{minipage} & \begin{minipage}[b]{\linewidth}\raggedright
0.264
\end{minipage} \\
\begin{minipage}[b]{\linewidth}\raggedright
2
\end{minipage} & \begin{minipage}[b]{\linewidth}\raggedright
Moderate NPDR
\end{minipage} & \begin{minipage}[b]{\linewidth}\raggedright
104
\end{minipage} & \begin{minipage}[b]{\linewidth}\raggedright
0.646
\end{minipage} & \begin{minipage}[b]{\linewidth}\raggedright
0.490
\end{minipage} & \begin{minipage}[b]{\linewidth}\raggedright
0.557
\end{minipage} \\
\begin{minipage}[b]{\linewidth}\raggedright
3
\end{minipage} & \begin{minipage}[b]{\linewidth}\raggedright
Severe NPDR
\end{minipage} & \begin{minipage}[b]{\linewidth}\raggedright
23
\end{minipage} & \begin{minipage}[b]{\linewidth}\raggedright
0.733
\end{minipage} & \begin{minipage}[b]{\linewidth}\raggedright
0.478
\end{minipage} & \begin{minipage}[b]{\linewidth}\raggedright
0.579
\end{minipage} \\
\begin{minipage}[b]{\linewidth}\raggedright
4
\end{minipage} & \begin{minipage}[b]{\linewidth}\raggedright
Proliferative DR
\end{minipage} & \begin{minipage}[b]{\linewidth}\raggedright
10
\end{minipage} & \begin{minipage}[b]{\linewidth}\raggedright
0.750
\end{minipage} & \begin{minipage}[b]{\linewidth}\raggedright
0.300
\end{minipage} & \begin{minipage}[b]{\linewidth}\raggedright
0.429
\end{minipage} \\
\midrule\noalign{}
\endhead
\bottomrule\noalign{}
\endlastfoot
\end{longtable}

\textbf{Table 6} Per-grade performance on the held-out Messidor-2 test set (n =
524).

\includegraphics[width=\textwidth]{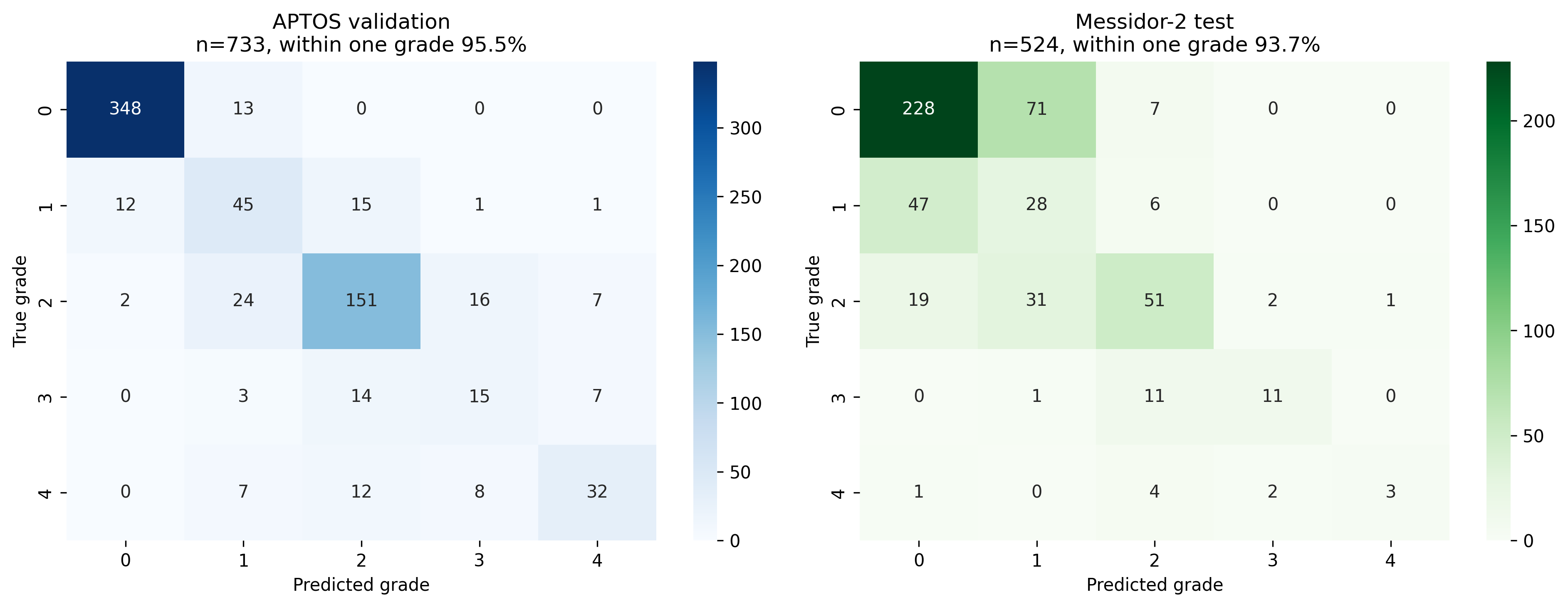}

\textbf{Fig. 5} Confusion matrices for the full model. Left: APTOS
validation (within one grade 95.5\%). Right: Messidor-2 held-out test
(within one grade 93.7\%). The dominant off-diagonal mass on Messidor-2
is grade-0 images assigned to grade 1, a one-step error, consistent with
preserved ordering but shifted decision boundaries under domain change

\subsection{Aggregate comparison and the generalization gap}

Table 7 places the two evaluations side by side. The QWK difference of
0.202 has a bootstrap 95\% confidence interval of 0.142-0.273 and
excludes zero in every one of 5,000 replicates, so the degradation is
statistically established. One contrast carries the paper's central
finding: accuracy falls by 19.3 percentage points and QWK by 0.202, yet
the within-one-grade rate falls only from 95.5\% to 93.7\%.

\begin{longtable}[]{@{}
  >{\raggedright\arraybackslash}p{(\columnwidth - 6\tabcolsep) * \real{0.2881}}
  >{\raggedright\arraybackslash}p{(\columnwidth - 6\tabcolsep) * \real{0.2548}}
  >{\raggedright\arraybackslash}p{(\columnwidth - 6\tabcolsep) * \real{0.2548}}
  >{\raggedright\arraybackslash}p{(\columnwidth - 6\tabcolsep) * \real{0.2023}}@{}}
\toprule\noalign{}
\begin{minipage}[b]{\linewidth}\raggedright
\textbf{Metric}
\end{minipage} & \begin{minipage}[b]{\linewidth}\raggedright
\textbf{APTOS validation (n = 733)}
\end{minipage} & \begin{minipage}[b]{\linewidth}\raggedright
\textbf{Messidor-2 test (n = 524)}
\end{minipage} & \begin{minipage}[b]{\linewidth}\raggedright
\textbf{Change}
\end{minipage} \\
\begin{minipage}[b]{\linewidth}\raggedright
Quadratic weighted kappa
\end{minipage} & \begin{minipage}[b]{\linewidth}\raggedright
0.882 {[}0.853, 0.906{]}
\end{minipage} & \begin{minipage}[b]{\linewidth}\raggedright
0.679 {[}0.613, 0.735{]}
\end{minipage} & \begin{minipage}[b]{\linewidth}\raggedright
−0.202
\end{minipage} \\
\begin{minipage}[b]{\linewidth}\raggedright
Accuracy
\end{minipage} & \begin{minipage}[b]{\linewidth}\raggedright
0.806 {[}0.778, 0.835{]}
\end{minipage} & \begin{minipage}[b]{\linewidth}\raggedright
0.613 {[}0.571, 0.655{]}
\end{minipage} & \begin{minipage}[b]{\linewidth}\raggedright
−0.193
\end{minipage} \\
\begin{minipage}[b]{\linewidth}\raggedright
Linear kappa
\end{minipage} & \begin{minipage}[b]{\linewidth}\raggedright
0.813
\end{minipage} & \begin{minipage}[b]{\linewidth}\raggedright
0.528
\end{minipage} & \begin{minipage}[b]{\linewidth}\raggedright
−0.285
\end{minipage} \\
\begin{minipage}[b]{\linewidth}\raggedright
Within one grade
\end{minipage} & \begin{minipage}[b]{\linewidth}\raggedright
95.5\%
\end{minipage} & \begin{minipage}[b]{\linewidth}\raggedright
93.7\%
\end{minipage} & \begin{minipage}[b]{\linewidth}\raggedright
−1.8 pp
\end{minipage} \\
\begin{minipage}[b]{\linewidth}\raggedright
Referable-DR sensitivity
\end{minipage} & \begin{minipage}[b]{\linewidth}\raggedright
0.879
\end{minipage} & \begin{minipage}[b]{\linewidth}\raggedright
0.620
\end{minipage} & \begin{minipage}[b]{\linewidth}\raggedright
−0.259
\end{minipage} \\
\begin{minipage}[b]{\linewidth}\raggedright
Referable-DR specificity
\end{minipage} & \begin{minipage}[b]{\linewidth}\raggedright
0.961
\end{minipage} & \begin{minipage}[b]{\linewidth}\raggedright
0.966
\end{minipage} & \begin{minipage}[b]{\linewidth}\raggedright
+0.005
\end{minipage} \\
\midrule\noalign{}
\endhead
\bottomrule\noalign{}
\endlastfoot
\end{longtable}

\textbf{Table 7} Aggregate performance with 95\% bootstrap confidence intervals
(5,000 replicates). Referable DR is defined as grade \textgreater= 2.

\subsection{Ablation study}

The ablation proceeds in two stages. Table 8 isolates each component
with a single training run under the protocol of Section 3.7, tracing
the path from a plain baseline to the full model. Because the
differences between the dual-branch variants proved small relative to
run-to-run variation, the three that matter for the fusion design were
then retrained under two further random seeds; Table 9 reports those
under mean and standard deviation.

\begin{longtable}[]{@{}
  >{\raggedright\arraybackslash}p{(\columnwidth - 6\tabcolsep) * \real{0.3399}}
  >{\raggedright\arraybackslash}p{(\columnwidth - 6\tabcolsep) * \real{0.2266}}
  >{\raggedright\arraybackslash}p{(\columnwidth - 6\tabcolsep) * \real{0.2167}}
  >{\raggedright\arraybackslash}p{(\columnwidth - 6\tabcolsep) * \real{0.2167}}@{}}
\toprule\noalign{}
\begin{minipage}[b]{\linewidth}\raggedright
\textbf{Variant}
\end{minipage} & \begin{minipage}[b]{\linewidth}\raggedright
\textbf{Isolates}
\end{minipage} & \begin{minipage}[b]{\linewidth}\raggedright
\textbf{APTOS QWK}
\end{minipage} & \begin{minipage}[b]{\linewidth}\raggedright
\textbf{Messidor-2 QWK}
\end{minipage} \\
\begin{minipage}[b]{\linewidth}\raggedright
EfficientNet-B0 + cross-entropy
\end{minipage} & \begin{minipage}[b]{\linewidth}\raggedright
Baseline
\end{minipage} & \begin{minipage}[b]{\linewidth}\raggedright
0.825
\end{minipage} & \begin{minipage}[b]{\linewidth}\raggedright
0.619
\end{minipage} \\
\begin{minipage}[b]{\linewidth}\raggedright
EfficientNet-B0 + ordinal head
\end{minipage} & \begin{minipage}[b]{\linewidth}\raggedright
Ordinal formulation
\end{minipage} & \begin{minipage}[b]{\linewidth}\raggedright
0.829
\end{minipage} & \begin{minipage}[b]{\linewidth}\raggedright
0.667
\end{minipage} \\
\begin{minipage}[b]{\linewidth}\raggedright
Dual-branch + concat (no gate)
\end{minipage} & \begin{minipage}[b]{\linewidth}\raggedright
Second backbone
\end{minipage} & \begin{minipage}[b]{\linewidth}\raggedright
0.868
\end{minipage} & \begin{minipage}[b]{\linewidth}\raggedright
0.707
\end{minipage} \\
\begin{minipage}[b]{\linewidth}\raggedright
Dual-branch, same preprocessing, + gate
\end{minipage} & \begin{minipage}[b]{\linewidth}\raggedright
Attention gate
\end{minipage} & \begin{minipage}[b]{\linewidth}\raggedright
0.844
\end{minipage} & \begin{minipage}[b]{\linewidth}\raggedright
0.693
\end{minipage} \\
\begin{minipage}[b]{\linewidth}\raggedright
\textbf{Full model (dual + gate + diverse preproc.)}
\end{minipage} & \begin{minipage}[b]{\linewidth}\raggedright
\textbf{Proposed}
\end{minipage} & \begin{minipage}[b]{\linewidth}\raggedright
\textbf{0.882}
\end{minipage} & \begin{minipage}[b]{\linewidth}\raggedright
\textbf{0.679}
\end{minipage} \\
\midrule\noalign{}
\endhead
\bottomrule\noalign{}
\endlastfoot
\end{longtable}

\textbf{Table 8} Component ablation, single seed (seed 42). Each row adds one
element to the row above it, except the fourth, which substitutes a
single shared preprocessing pipeline for the two distinct ones to
isolate the effect of preprocessing diversity.

Read down the Messidor-2 column, two components carry clear and
consistent weight. The ordinal head lifts held-out kappa from 0.619 to
0.667, and adding the second backbone lifts it again to 0.707. Both
gains hold up across seeds (below) and are the substance of the
architecture. The last two rows are where the single-seed reading is
misleading: the full model, with its attention gate and two
preprocessing pipelines, scores 0.679, below the simpler no-gate variant
at 0.707. Taken alone, that table would suggest the gate and
preprocessing diversity actively harm generalization. They do not, and
the reason they appear to is instructive.

The three dual-branch variants differ by amounts comparable to the
run-to-run variation of a single configuration, so a one-seed comparison
cannot separate a real effect from initialisation noise. Retraining them
under three seeds in total resolves the question. Table 9 reports the
outcome.

\begin{longtable}[]{@{}
  >{\raggedright\arraybackslash}p{(\columnwidth - 4\tabcolsep) * \real{0.3767}}
  >{\raggedright\arraybackslash}p{(\columnwidth - 4\tabcolsep) * \real{0.2936}}
  >{\raggedright\arraybackslash}p{(\columnwidth - 4\tabcolsep) * \real{0.3297}}@{}}
\toprule\noalign{}
\begin{minipage}[b]{\linewidth}\raggedright
\textbf{Variant}
\end{minipage} & \begin{minipage}[b]{\linewidth}\raggedright
\textbf{APTOS QWK (mean ± sd)}
\end{minipage} & \begin{minipage}[b]{\linewidth}\raggedright
\textbf{Messidor-2 QWK (mean ± sd)}
\end{minipage} \\
\begin{minipage}[b]{\linewidth}\raggedright
Dual-branch + concat (no gate)
\end{minipage} & \begin{minipage}[b]{\linewidth}\raggedright
0.873 ± 0.004
\end{minipage} & \begin{minipage}[b]{\linewidth}\raggedright
0.694 ± 0.018
\end{minipage} \\
\begin{minipage}[b]{\linewidth}\raggedright
Dual-branch, same preprocessing, + gate
\end{minipage} & \begin{minipage}[b]{\linewidth}\raggedright
0.848 ± 0.006
\end{minipage} & \begin{minipage}[b]{\linewidth}\raggedright
0.660 ± 0.033
\end{minipage} \\
\begin{minipage}[b]{\linewidth}\raggedright
\textbf{Full model (dual + gate + diverse preproc.)}
\end{minipage} & \begin{minipage}[b]{\linewidth}\raggedright
\subsection{± 0.008}
\end{minipage} & \begin{minipage}[b]{\linewidth}\raggedright
\subsection{± 0.021}
\end{minipage} \\
\midrule\noalign{}
\endhead
\bottomrule\noalign{}
\endlastfoot
\end{longtable}

\textbf{Table 9} The three dual-branch variants over three random seeds (42, 1,
2), mean ± sample standard deviation. Seeds 1 and 2 were trained for 8
epochs, seed 42 for 10; validation kappa had plateaued by epoch 4 in all
runs, and the seed-42 dual-concat value differed from its 8-epoch
counterpart by 0.006.

The no-gate variant and the full model are now separated by 0.005 in
mean held-out kappa (0.694 against 0.689), against a combined standard
deviation of about 0.04. They are statistically indistinguishable. The
apparent 0.028 advantage of plain concatenation in Table 8 was an
artefact of a single seed: the full model recorded its weakest run at
seed 42 and its strongest at seed 2 (0.713), and the ranking of the two
variants swapped between seeds. Neither the attention gate nor the
second preprocessing pipeline produces a measurable change in held-out
grading, in either direction.

One further pattern in Table 9 is worth stating, because it reinforces
the theme of the paper. The in-domain standard deviations are 0.004 to
0.008, while the held-out standard deviations are 0.018 to 0.033 , three to five times larger. Domain shift does not merely lower
performance; it makes performance less stable from run to run. A single
held-out number, of the kind most DR papers would report, therefore
carries substantially more uncertainty than a single in-domain number,
which is a further reason to prefer multi-seed reporting for any
cross-domain claim.

\subsection{Summary}

The model grades DR with QWK 0.882 in-domain and 0.679 on a held-out multi-centre test set for this run (0.689 $\pm$ 0.021 across three random seeds). The 0.202 gap is significant. The ordinal head
keeps 93.7\% of even the degraded predictions within one grade of
reference. Referable-DR sensitivity falls from 0.879 to 0.620 while
specificity holds above 0.96.

\section{Discussion}

\subsection{Ordering survives domain shift; grade boundaries do not}

The most informative result in this study is a dissociation. Under
domain shift, accuracy falls 19.3 points and linear kappa falls 0.285,
but the within-one-grade rate falls only 1.8 points, from 95.5\% to
93.7\%. The model does not lose its ordering of disease severity when
the imaging domain changes; it loses its calibration of where the
boundaries between grades lie. The learned representation still ranks a
proliferative retina above a mild one, errors remain local, but
the decision boundaries, fitted on one intensity distribution, sit in
the wrong place on another.

The confusion structure of Table 6 and Figure 5 supports this reading
directly. The dominant Messidor-2 error is 71 grade-0 images assigned to
grade 1: a systematic one-step shift at the lowest decision threshold,
not a scattering of predictions. A model whose features had genuinely
failed to transfer would produce diffuse, high-distance errors, and a
93.7\% within-one-grade rate rules that out.

The distinction matters practically, because the two diagnoses imply
different remedies. If the representation had collapsed, the appropriate
response would be adversarial domain adaptation or retraining. If the
ordering is intact and only the boundaries have moved, then a light
recalibration of the head on a small labelled sample from the target
domain, with the backbones frozen, should recover a substantial
fraction of the gap at negligible cost. This work states that as a
hypothesis and does not test it. One methodological caveat applies:
because the four binary subtasks carry independent weight vectors
(Section 3.6), threshold placement is not isolated in a small set of
bias parameters, so the recalibration would touch the whole head.
Adopting the CORAL constraint of Cao et al. [28], under which the
subtasks share a single weight vector and differ only in four bias
terms, would both restore the rank-monotonicity guarantee and localise
threshold placement to four parameters, making the hypothesis directly
testable. That substitution is a natural next step.

\subsection{Preprocessing diversity and attention-gated fusion}

The ablation is unambiguous about which parts of the design earn their
place. Adding the second backbone is the single largest architectural
gain on held-out data, and the ordinal head contributes a clear
improvement over cross-entropy; both survive multi-seed testing. The
attention gate and the pairing of two distinct preprocessing pipelines
do not. Over three seeds the full model and a plain-concatenation,
single-preprocessing variant are separated by 0.005 in mean held-out
kappa against a combined standard deviation near 0.04, which is no
separation at all. We therefore make no claim that the gate or
preprocessing diversity improve grading, and equally none that they harm
it; the honest statement is that they leave held-out performance
unchanged within the resolution of the experiment. They remain
reasonable architectural choices, the gate is a principled way to
combine two streams and adds negligible parameters, but they are
choices, not sources of measured gain, and the paper is written to say
exactly that.

Two cautions attach to this reading. The truncation of the B3 feature
vector described in Section 3.5 discards 16.7\% of that branch, so the
gate is being asked to arbitrate over an impaired input; a learned
projection might change the picture, and we did not test it. And the
single-seed reversal in Table 8, where plain concatenation appeared to
beat the full model outright, is a concrete illustration of how a
one-run ablation can manufacture a contribution that does not exist , in this case a negative one. The episode is the reason the surviving
claims in this paper are restricted to those that held across seeds.

\subsection{Comparison with existing research}

Direct numerical comparison against the studies in Table 1 is not
meaningful, because they differ in datasets, grade schemes, splits, and
metrics. This is the substantive point rather than a rhetorical evasion:
the reported accuracies of 92-99\% in Section 2.1 were obtained
in-domain, and the present work shows that an in-domain QWK of 0.882
coexists with a held-out QWK of 0.679 (0.689 $\pm$ 0.021 across three seeds) in the same model. In-domain
figures from different protocols therefore cannot be ranked against one
another, and a system reporting 98\% accuracy in-domain has not thereby
been shown to outperform one reporting 80.6\%.

The claim of this paper is correspondingly narrow. We do not claim
state-of-the-art in-domain accuracy; we claim a measured,
uncertainty-quantified account of what a competitive DR architecture
does when the imaging domain changes, and an empirical characterisation
of the form its failure takes.

\subsection{Clinical implications}

Referable-DR sensitivity of 0.620 on the held-out set, against 0.879
in-domain, is the number with clinical consequence, and it should be
read as disqualifying for autonomous screening triage in its present
form. A useful benchmark is the standard adopted by the UK National
Health Service Diabetic Eye Screening Programme, following the British
Diabetes Association consensus, which requires a screening test to
demonstrate at least 80\% sensitivity and at least 95\% specificity for
referable disease [37, 38]. The model meets
the specificity requirement comfortably in both domains (0.961 and
0.966) but falls well short of the 80\% sensitivity floor on held-out
data at 62.0\%, having met it in-domain at 87.9\%. A screening
instrument that misses roughly two of every five referable cases in a
new imaging environment cannot be deployed as a first-line filter,
whatever its kappa. That the failure is one of missed disease rather
than false alarm, specificity holds while sensitivity collapses , is the less acceptable of the two directions for a screening tool, since
the cost of a missed referral is borne by the patient.

\subsection{Limitations}

\begin{itemize}
\item
  The multi-seed ablation covers three seeds; the attention gate and
  preprocessing diversity are shown to be within noise of simpler
  variants rather than proven equivalent, and a larger seed budget could
  narrow the interval further. Seeds 1 and 2 were trained for 8 epochs
  against 10 for seed 42, a minor inconsistency on a plateaued
  validation curve.
\item
  Messidor-2 training images were included in the training set, so the
  Messidor-2 test evaluation measures held-out rather than zero-shot
  performance. The true cross-dataset penalty is therefore likely larger
  than the 0.202 reported here, and this figure should be read as a
  lower bound.
\item
  The fusion stage truncates the EfficientNet-B3 feature vector to the
  width of the B0 vector, discarding 16.7\% of the B3 channels and
  assuming an unlearned channel correspondence between the two
  backbones. A learned projection into a shared space would be the
  principled alternative.
\item
  The ordinal head does not enforce rank monotonicity, so
  rank-inconsistent probability patterns are representable and are not
  detected by the threshold-counting decision rule. The observed
  within-one-grade rates are empirical, not guaranteed.
\item
  Only two public datasets were used, both retrospectively collected.
  Prospective multi-centre clinical validation is required before any
  deployment claim.
\item
  Supervision is image-level; no lesion-level annotation was available,
  so the model cannot localise the pathology driving each grade.
\item
  Grades 3 and 4 have supports of 39 and 59 images in the APTOS split
  and 23 and 10 in the Messidor-2 test set, so per-grade estimates at
  the severe end are imprecise.
\item
  Domain alignment relies on histogram matching against a single
  arbitrarily chosen reference image; adversarial or feature-space
  adaptation was not attempted.
\item
  Only the dataset splits were seeded. Weight initialisation, sampler
  draws, and augmentation were not, so exact numerical reproduction of
  the reported figures is not possible from the released code alone.
\end{itemize}

\section{Conclusions}

This work presented a dual-resolution, attention-gated DR grading
framework with an ordinal regression head, and evaluated it with
explicit attention to what happens when the imaging domain changes. The
model achieved QWK 0.882 (95\% CI 0.853-0.906) in-domain on APTOS and
0.679 (95\% CI 0.613-0.735; 0.689 $\pm$ 0.021 across three seeds) on a held-out multi-centre Messidor-2 test
set, a gap of 0.202 (95\% CI 0.142-0.273) that is statistically
significant and persists despite histogram matching.

The principal finding is not the gap itself but its structure. Ordering
survives domain shift while grade-boundary placement does not: 93.7\% of
predictions remain within one grade of reference even as accuracy falls
19.3 points, and the dominant error is a systematic one-step shift at
the lowest threshold. This points to recalibration of the decision head,
rather than adaptation of the representation, as the cheaper route to
recovering the loss, a prediction this work states but does not test.

The immediate implication for the field is methodological. A model
reporting 88\% agreement in-domain delivered 62\% referable-DR
sensitivity on images from different cameras. In-domain metrics,
reported without external evaluation and without uncertainty, do not
characterise a screening system. Future work should complete the
component ablation, adopt a rank-consistent ordinal head and test the
recalibration hypothesis, replace feature truncation with a learned
projection, retrain under a strict APTOS-only protocol to measure the
true zero-shot penalty, and validate prospectively across clinical
centres.

\textbf{Funding}

This research received no external funding.

\textbf{Conflicts of Interest}

The author declares no conflict of interest.

\textbf{Data and Code Availability}

The datasets used are publicly available: APTOS 2019
(https://www.kaggle.com/c/aptos2019-blindness-detection) and Messidor-2
(https://www.adcis.net/en/third-party/messidor2/). The complete training and evaluation code, the self-contained reproduction notebook, the split manifests, and the figures are available at \url{https://github.com/Dr-Afshan/DR_Grading} under the MIT license. An archived, version-tagged release is deposited at Zenodo (DOI: \texttt{10.5281/zenodo.21739226}). The unit tests reproduce the reported metrics from the confusion matrices without a GPU or the datasets.

\textbf{Abbreviations}

AMP: Automatic Mixed Precision; CLAHE: Contrast-Limited Adaptive
Histogram Equalisation; CNN: Convolutional Neural Network; CORAL , COnsistent RAnk Logits; DR: Diabetic Retinopathy; NPDR , Non-Proliferative Diabetic Retinopathy; OR-CNN: Ordinal Regression
Convolutional Neural Network; PDR: Proliferative Diabetic
Retinopathy; QWK: Quadratic Weighted Kappa; SE , Squeeze-and-Excitation.

\section*{Declarations}

\noindent\textbf{Funding:} No funding was received to assist with the preparation of this manuscript.

\noindent\textbf{Competing interests:} The author has no competing interests to declare that are relevant to the content of this article.

\noindent\textbf{Ethics approval:} This study used two publicly available, fully de-identified retinal image datasets (APTOS 2019 and Messidor-2). No new data were collected from human participants by the author, and no identifiable personal information was accessed. Ethical approval was therefore not required for this secondary analysis of anonymised public data; the original datasets were collected under their respective ethical frameworks.

\noindent\textbf{Consent to participate:} Not applicable. The study used existing anonymised public datasets and did not involve recruitment of human participants by the author.

\noindent\textbf{Data and code availability:} The datasets analysed are publicly available: APTOS 2019 (\url{https://www.kaggle.com/c/aptos2019-blindness-detection}) and Messidor-2 (\url{https://www.adcis.net/en/third-party/messidor2/}). The complete training and evaluation code, the self-contained reproduction notebook, and the split manifests are available at \url{https://github.com/Dr-Afshan/DR_Grading} under the MIT license, with an archived version deposited at Zenodo (DOI: 10.5281/zenodo.21739226).

\noindent\textbf{Author contributions:} A.H. conceived and designed the study, implemented the models and experiments, analysed the results, and wrote the manuscript.

\section*{References}
\begin{enumerate}[label={[\arabic*]}, leftmargin=*, itemsep=1pt, topsep=2pt]
\item Siddharth, M.; Suguna, S.; Mallika, K.; Pandiri, A.; et al. Predict diabetic retinopathy in early stages: a novel ensemble model using EfficientNets and an automated system to detect the disease. Int. J. Innov. Technol. Explor. Eng. 2022, 11, 38–48.
\item Bhavya, M. Detection of diabetic retinopathy using deep learning. Int. J. Sci. Technol. Eng. 2022, 10, 1880–1886.
\item Pradeepa, R.; Mohan, V. Epidemiology of type 2 diabetes in India. Indian J. Ophthalmol. 2021, 69, 2932.
\item Saeedi, P.; Petersohn, I.; Salpea, P.; et al. Global and regional diabetes prevalence estimates for 2019 and projections for 2030 and 2045: results from the IDF Diabetes Atlas. Diabetes Res. Clin. Pract. 2019, 157, 107843.
\item IDF Diabetes Atlas, 9th ed. Available online: https://diabetesatlas.org/atlas/ninth-edition (accessed 2 October 2023).
\item Diabetes. Pan American Health Organization. Available online: https://www.paho.org/en/topics/diabetes (accessed 10 November 2023).
\item Memari, N.; Abdollahi, S.; Ganzagh, M.; Moghbel, M. Computer-assisted diagnosis (CAD) system for diabetic retinopathy screening using color fundus images using deep learning. In Proceedings of the IEEE Student Conference on Research and Development (SCOReD), 2020; pp. 69–73.
\item Gulshan, V.; Peng, L.; Coram, M.; et al. Development and validation of a deep learning algorithm for detection of diabetic retinopathy in retinal fundus photographs. JAMA 2016, 316, 2402–2410.
\item Karthik, S.; Geetha, M.; Prabhavathi, K.; et al. Early detection and severity classification of diabetic retinopathy using convolutional neural networks. SN Comput. Sci. 2025, 6, 819.
\item Bappi, M.; Juthy, J.; Kim, K. Deep learning-based diabetic retinopathy recognition and grading: challenges, gaps, and an improved approach, a survey. ICT Express 2025, 11, 993–1013.
\item Tan, M.; Le, Q. EfficientNet: rethinking model scaling for convolutional neural networks. In Proceedings of the International Conference on Machine Learning (ICML), Long Beach, CA, USA, 2019; pp. 6105–6114.
\item Mohanty, C.; Acharya, B.; Kokkoras, F.; et al. Using deep learning architectures for detection and classification of diabetic retinopathy. Sensors 2023, 23, 5726.
\item Raiaan, M.; Mukta, M.; Fatema, K.; et al. A lightweight robust deep learning model gained high accuracy in classifying a wide range of diabetic retinopathy images. IEEE Access 2023, 11, 42361–42388.
\item Raman, D.; Nishanthi, S. Diagnosis of diabetic retinopathy by using EfficientNet-B7 CNN architecture in deep learning. In Proceedings of the International Conference on Computer Science and Software Studies (ICSCSS), Chennai, India, 2023; pp. 430–435.
\item Nivedha, A.; Sreelakshmi, T. Detection and classification of diabetic retinopathy using deep learning. Int. J. Sci. Technol. Eng. 2023, 11, 856–860.
\item Mondal, S.; Mandal, N.; Singh, K.; Singh, A.; Izonin, I. EDLDR: an ensemble deep learning technique for detection and classification of diabetic retinopathy. Diagnostics 2023, 13, 124.
\item Nazir, T.; Nawaz, M.; Rashid, J.; et al. Detection of diabetic eye disease from retinal images using a deep learning based CenterNet model. Sensors 2021, 21, 5283.
\item Subbanna, M. Diabetic retinopathy classification based on fundus image using convolutional neural network (CNN) with MobileNetV2. In Advances in Computing and Data Sciences; Springer: Singapore, 2023; pp. 89–102.
\item Shanthala, K. V.; Kundur, N. C. DR-EfficientNet-L: a distributed deep learning architecture for efficient detection and grading of diabetic retinopathy. Eng. Technol. Appl. Sci. Res. 2025, 15, 28362-28367. https://doi.org/10.48084/etasr.13201
\item Rahman, M.; et al. RSG-Net: a deep learning based model for diabetic retinopathy grading. Sci. Rep. 2025, 15, 3763. https://doi.org/10.1038/s41598-025-87171-9
\item Deep learning models for grading diabetic retinopathy: a systematic review and meta-analysis. Front. Endocrinol. 2026. https://doi.org/10.3389/fendo.2026.1853785
\item Fu, Y.; Ju, Y.; Zhang, D. MSEF-Net: a multi-scale EfficientNet fusion network for diabetic retinopathy grading. Biomed. Signal Process. Control 2024, 92, 106168.
\item Shakibania, H.; Raoufi, S.; Pourafzal, A.; et al. Dual-branch deep learning network for detection and stage grading of diabetic retinopathy. Biomed. Signal Process. Control 2024, 90, 105845.
\item Romero-Oraa, R.; Garcia, M.; Oraa-Perez, J.; Lopez, M.; Hornero, R. Attention-based deep learning framework for automated diabetic retinopathy grading. Comput. Methods Programs Biomed. 2024, 240, 107688.
\item Zhang, W.; Chen, Y.; Li, X.; Wang, Z.; Liu, H. Attention-guided deep learning for interpretable diabetic retinopathy analysis. Computers 2025, 14, 187.
\item Hu, J.; Shen, L.; Sun, G. Squeeze-and-Excitation Networks. In Proceedings of the IEEE/CVF Conference on Computer Vision and Pattern Recognition (CVPR), Salt Lake City, UT, USA, 2018; pp. 7132–7141.
\item Niu, Z.; Zhou, M.; Wang, L.; Gao, X.; Hua, G. Ordinal regression with multiple output CNN for age estimation. In Proceedings of the IEEE Conference on Computer Vision and Pattern Recognition (CVPR), Las Vegas, NV, USA, 2016; pp. 4920–4928.
\item Cao, W.; Mirjalili, V.; Raschka, S. Rank consistent ordinal regression for neural networks with application to age estimation. Pattern Recognit. Lett. 2020, 140, 325–331.
\item Shi, X.; Cao, W.; Raschka, S. Deep neural networks for rank-consistent ordinal regression based on conditional probabilities. Pattern Recognit. Lett. 2023, 168, 112–120.
\item Chilukoti, S.; Maida, A.; Hei, X. Reliable diabetic retinopathy grading using ordinal learning and quadratic weighted kappa optimization. BMC Med. Inform. Decis. Mak. 2024, 24, 112.
\item Voets, M.; Møllersen, K.; Bongo, L. Reproduction study using public data of: development and validation of a deep learning algorithm for detection of diabetic retinopathy in retinal fundus photographs. PLoS ONE 2019, 14, e0217541.
\item APTOS 2019 blindness detection dataset. Available online: https://www.kaggle.com/c/aptos2019-blindness-detection (accessed 10 November 2023).
\item Decencière, E.; Zhang, X.; Cazuguel, G.; et al. Feedback on a publicly distributed image database: the Messidor database. Image Anal. Stereol. 2014, 33, 231–234.
\item Graham, B. Kaggle Diabetic Retinopathy Detection competition report, 2015. Available online: https://www.kaggle.com/c/diabetic-retinopathy-detection/discussion/15801 (accessed 10 November 2023).
\item Zuiderveld, K. Contrast limited adaptive histogram equalization. In Graphics Gems IV; Heckbert, P., Ed.; Academic Press: San Diego, CA, USA, 1994; pp. 474–485.
\item Gonzalez, R.; Woods, R. Digital Image Processing, 4th ed.; Pearson: New York, NY, USA, 2018.
\item NHS Diabetic Eye Screening Programme. Diabetic eye screening: commission and provide. NHS England / UK Health Security Agency, 2024. Available online: https://www.gov.uk/government/publications/diabetic-eye-screening-commission-and-provide (accessed 2024).
\item Scanlon, P. H. The English National Screening Programme for diabetic retinopathy 2003-2016. Acta Diabetol. 2017, 54, 515–525.
\end{enumerate}
\end{document}